\documentclass[letterpaper, 10pt, conference]{ieeeconf}
\IEEEoverridecommandlockouts    
\usepackage{multirow}
\usepackage{booktabs}
\usepackage{lipsum}
\usepackage{amsmath}
\usepackage{amssymb}
\usepackage[ruled,vlined]{algorithm2e}
\usepackage{graphicx}
\usepackage{url}

\graphicspath{{./Figures/}}

\title{\LARGE \bf BLInD: Learning Driver Intent as a Distribution over Future Ego Trajectories}

\author{Flavian Pegado, Ronit Hire, Shreyas Rajesh, and Soham Phade%
\thanks{All authors are with Wayve Technologies. Emails:
{\tt\small \{flavian.pegado, ronit.hire, shreyas.rajesh, soham.phade\}@wayve.ai}}%
}

\begin{document}
	\maketitle
	\thispagestyle{empty}
	\pagestyle{empty}
	
	\begin{abstract}
    We present BLInD (Blind Learned Intent Distribution), a compact network that maps recent vehicle-state history (e.g. speed, curvature, indicator, and vehicle type) to a top-k distribution of future ego trajectories, with no camera, LiDAR, map, or object-track inputs. We find that vehicle-state history alone is sufficient to learn a useful multimodal distribution over near-term ego trajectories, and its low-latency nature makes it well-suited for safety-critical deployment. We investigate two distribution architectures, autoregressive (AR) and flow-matching, and train on both mixed-platform open-source and Wayve datasets. Both generalize without dataset-specific adaptation; the flow-matching model achieves best top-k ADE/FDE of 0.15/0.37 m on Wayve, 0.15/0.36 m on Waymo, and 0.28/0.59 m on nuScenes, with the AR model reaching comparable coverage. Integrating the distributions into an AEB trigger task, a strict all-candidates policy reduces false positives from 1.51\% to 0.11\% with AR (13.7× reduction, 94.9\% TP) and to 0.06\% with flow-matching (25.1× reduction, 98.7\% TP) compared to a 1-CTRV policy with 100\% true positive score. BLInD runs in 0.87 ms with the AR head and 2.9 ms with the flow-matching head on an NVIDIA DRIVE Orin ECU making it compatible with real-time deployment on automotive ECUs. While existing learned distribution models rely on scene context and blind vehicle-state models typically collapse to a single path, BLInD is learned, blind, and cross-domain simultaneously, a combination not demonstrated by prior work. These results show that such a distribution provides a controllable and plausible intent sampling interface for downstream systems, with AEB as one instantiation.
	\end{abstract}
	
	\section{Introduction}
	\label{sec:introduction}
    
    In the context of Advanced Driver Assistance Systems (ADAS), safe collaborative driving requires anticipating what the human driver is most likely to do next. Traditional systems often rely on a single deterministic extrapolation, like the constant-turn-rate-and-velocity (CTRV) model \cite{ctrv_model}, or a blend of heuristic trajectories \cite{houenou2013}.

    However, near-term driver intent, as expressed through the ego trajectory over the next few seconds, is inherently multimodal.
    A learned distribution over plausible futures strikes a practical middle ground; more expressive than a single path, more focused than every feasible trajectory, unlocking scaling to longer horizons, and lets downstream applications query that uncertainty directly.

    We study a deliberately blind realization of this idea: a compact network that receives only recent speed, curvature, indicator state, and vehicle type, with no camera images, object tracks, or map inputs. Rather than replacing perception-rich planning, it acts as a lightweight ego-dynamics prior that can be trained across platforms and directly queried by downstream applications.

    This paper makes three contributions. First, we introduce BLInD, a blind ego-intent predictor that is simultaneously learned, distributional, and cross-domain, a combination not demonstrated by prior work. Second, we evaluate two distribution architectures, autoregressive and flow-matching, showing both generalize across Wayve, Waymo, and nuScenes without dataset-specific adaptation. Third, we validate downstream utility via Automatic Emergency Braking (AEB): the predicted distribution supports a controllable TP/FP operating curve, where a strict $all$-candidates collision check with drivable area boundary query reduces false positives by over 25× relative to CTRV while preserving most true-positive behavior. AEB is just one instantiation of a broader interface applicable wherever downstream logic today relies on a single deterministic ego trajectory.
    
    \begin{figure*}
        \centering
        \includegraphics[width=0.7\linewidth, clip]{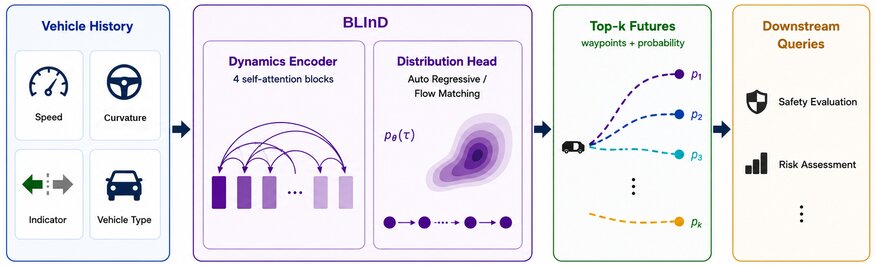}
        \caption{BLInD takes recent vehicle-history signals, encodes them with a compact dynamics encoder, and uses either an autoregressive or flow-matching distribution head to produce top-k future ego trajectories with probabilities. Downstream modules can query this distribution  for tasks such as safety evaluation and risk assessment.}
        \label{fig:interface}
    \end{figure*}
	

    \section{Related Work}
    \label{sec:related_work}

    \textbf{Behavioral intent prediction.} Intent prediction for driving is often framed as estimating maneuvers, future motion, and risk for intelligent vehicles~\cite{intent_survey}. That framing is useful for interaction understanding, but it is too coarse for short-horizon ego safety decisions. In AEB-like settings, the question is not only whether the driver is changing lanes; it is whether the set of plausible future ego footprints collides with a safety boundary. We therefore model intent directly as a continuous distribution over ego trajectories.
    
    \textbf{Multimodal trajectory forecasting.} Methods such as MTP, CoverNet, MultiPath, and Trajectron++ establish that future motion is inherently multi-valued~\cite{mtp,covernet,multipath,trajectron}, and benchmarks such as nuScenes, Waymo Open Motion Dataset, and Argoverse standardize evaluation via top-$k$ metrics including minADE, minFDE, and miss rate~\cite{nuscenes,waymo,argoverse}. Our work borrows this distributional view but changes the input and target: these systems predict other agents using perception, maps, and interaction context, whereas we predict the ego driver's future from blind vehicle dynamics alone.

    \textbf{Cross-domain forecasting and physics baselines.} Cross-domain methods demonstrate that learned distributional models can generalize across datasets~\cite{unitraj}, but rely on HD maps and agent context. Classical models such as CTRV~\cite{ctrv_model} are portable and competitive for short-horizon ego motion but single-mode; they cannot distinguish a collision implied by every plausible intent from one implied by a single extrapolated path.

    \textbf{Blind ego-dynamic prediction.} Recent work shows that learned ego-status models match perception-rich planners on open-loop benchmarks~\cite{admlp,bevplanner}, confirming that vehicle dynamics carry a strong predictive signal. However, these models collapse to a single path and target planning metrics, not downstream applications that require reasoning over a distribution of plausible futures. BLInD combines what each line leaves open: a blind, distributional, and cross-domain ego-intent predictor.
    
    \textbf{Prediction-aware safety.} Safety and reachability work connects prediction quality to downstream decisions by treating predicted trajectories as forward reachable sets~\cite{reachability}. Our AEB evaluation follows the same principle: the top-$k$ futures form an intent-conditioned reachable set, and the trigger policy asks whether one, some, or all plausible futures collide.
	
	\section{Method}
    \label{sec:method}

    \subsection{Blind Driver-Intent Interface}
    
    The model consumes vehicle-state history $\mathcal{H}$ consisting of speed, rear-axle curvature, vehicle type, indicator states and emits a trajectory distribution $p(\tau_{1:K} \mid \mathcal{H})$ where each $\tau_k$ is a sequence of $T=10$ future waypoints 0.2 secs apart. We deliberately choose low bandwidth signals as inputs and avoid scene information from image tokens, object tracks, HD maps, route maps or lane features. We propose sampling from this learned distribution and exposing top-$k$ trajectories based on their likelihoods as an interface for downstream tasks and later show the use of such an interface for an AEB system as a potential application.

    \subsection{Compact Network Design}
    
    The network to learn this distribution is intentionally small.
    The dynamics encoder embeds each vehicle state variable in a 64-dimensional space and applies four lightweight self-attention blocks with four heads as shown in Fig.~\ref{fig:interface}. On top of this shared encoder, we explore an autoregressive head over speed and curvature sequences and a flow-matching head over trajectory samples for modeling driver intent. This compact encoder and lightweight decoder design enables deployment latencies of 0.87 ms for the AR head and 2.9 ms for the flow-matching head on an NVIDIA DRIVE Orin ECU \cite{nvidia_drive_orin}.

    \subsection{Autoregressive Sampling}

    The autoregressive head (AR) models driver intent as a mode-conditioned distribution over future speeds and curvatures. Following autoregressive sequence modeling~\cite{autoregressive_rnn}, it first predicts a categorical distribution over $\mathcal{M}$ high-level intent modes $(z_{1:\mathcal{M}})$, and then uses a lightweight GRU decoder~\cite{gru} for each mode. At each future step, the decoder uses the previous speed and curvature bins and predicts categorical logits for the next speed bin and curvature logits conditioned on that speed. The decoded speed-curvature sequences are then integrated into ego-frame waypoints as likely driver trajectories.
    
    We represent speed using 64 uniformly spaced bins over \([0,40]\) m/s and curvature is represented using 256 non-uniform bins over \([-0.25,0.25]\) m\(^{-1}\), with finer resolution near zero curvature and coarser resolution near the extremes. During training, the decoder is trained by a teacher using the ground-truth speed and curvature bins from the previous step and trained with a cross-entropy (CE) loss. We decode all $\mathcal{M}$ intent modes and select the mode with the lowest trajectory reconstruction loss and only propagate the speed and curvature losses corresponding to that mode. We also use the selected mode as a one-hot label for an auxiliary mode classification CE loss.
    
    At inference, top-$k$ modes from \(p_\theta(z_{1:\mathcal{M}}\mid \mathcal{H})\) are selected and for each selected mode, we greedily decode speed and curvature bins up to $T$ steps. This select and decode strategy gives us a ranked set of plausible future trajectories for downstream tasks.

    \subsection{Flow-Matching Trajectory Sampling}
    
    The flow-matching head models driver intent as a continuous distribution over future ego-frame waypoints, following recent flow-matching and rectified-flow generative modeling work~\cite{flow_matching,rectified_flow}. Given the encoded vehicle history, it learns a conditional velocity field over trajectory states rather than discretizing speed and curvature. The decoder is lightweight: the 10 noisy future waypoints are projected into 64-dimensional tokens, passed through two cross-attention/self-attention blocks conditioned on the vehicle-history tokens, and projected back to 2D velocity updates.
    
    During training, the ground-truth future trajectory is treated as the clean data sample. For each training example, we draw a base trajectory state from Gaussian noise and sample a random interpolation time between zero and one. This gives an intermediate noisy trajectory on the straight path between the Gaussian base state and the ground-truth trajectory. The model is trained to predict the constant rectified-flow velocity from the base state to the ground-truth trajectory, using an MSE loss between the predicted and target velocities. We sample multiple time/noise pairs per trajectory, so each ground-truth future supervises the flow field at several points from noise to data.
    
    At inference, we generate the top-1 trajectory by integrating from a zero base state, making it deterministic for a fixed vehicle history. Additional $k-1$ candidates are generated by integrating from Gaussian base states, giving stochastic alternatives from the same conditional flow. All candidates use 10 Euler steps of the learned velocity field. Since the flow sampler does not rank candidates by likelihood, downstream queries are assigned  uniform weights over the generated trajectories.

    \begin{figure*}[t]
    \centering
    \includegraphics[width=0.27\textwidth,height=0.12\textheight,keepaspectratio,clip]{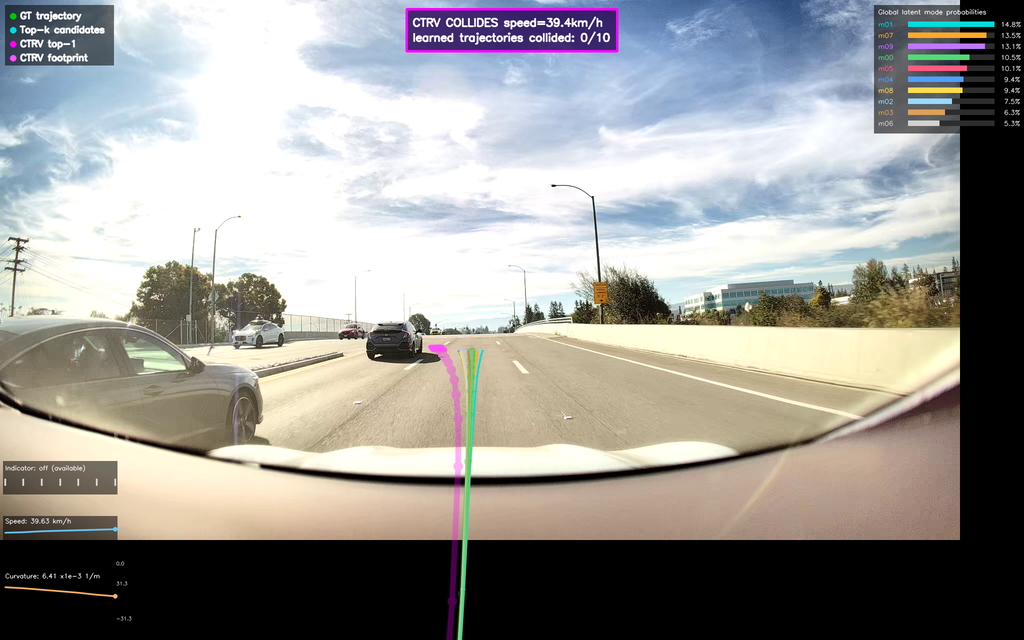}
    \includegraphics[width=0.27\textwidth,height=0.12\textheight,keepaspectratio,clip]{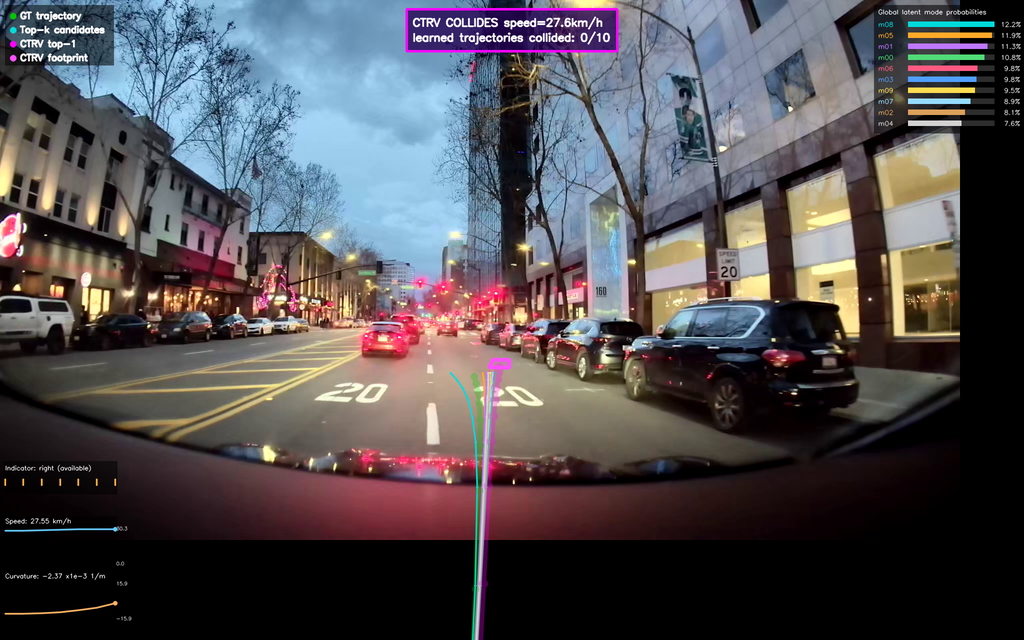}
    \includegraphics[width=0.27\textwidth,height=0.12\textheight,keepaspectratio,clip]{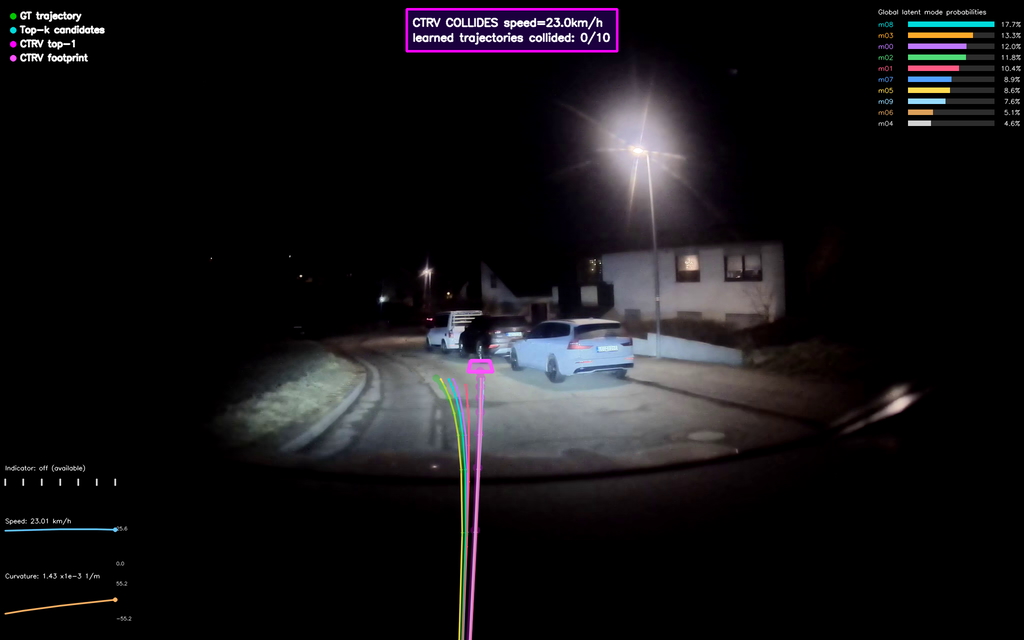}

    \caption{A typical CTRV extrapolation extends beyond the drivable safe space and predicts a potential collision (ego box at the final waypoint intersecting with objects in the scene), while the learned top-$k$ ego futures show plausible collision avoidance behaviors. $k$-CTRV rollouts can similarly produce collision-free trajectories but they end up spanning the entire feasible space failing to capture any driver intent leading to poor TP scores in an AEB setting.}
    \label{fig:ar_hook}
\end{figure*}
    
    \section{Evaluation}
    \label{sec:evaluation}
    
    Our evaluation covers two aspects of the learned distribution: coverage quality and downstream utility.
    
    \subsection{Open-Loop Intent Distribution Evaluation}
    
    Open-loop intent evaluation measures whether the predicted future distribution covers the realized driver trajectory. Given the recent vehicle history, each method produces a ranked set of $k$ candidate ego futures, which we compare against the observed future. We report top-$k$ average displacement error (ADE), final displacement error (FDE), and miss rate defined as how often no candidate ends within the specified distance threshold, 2 m here. For top-$k$, ADE and FDE use the candidate with the lowest trajectory error. This evaluates coverage of the predicted trajectory set, not only accuracy of the first-ranked trajectory. We evaluate on three validation sources: the Wayve split, an internal driving dataset measuring performance in the primary development domain, and the open-source Waymo and nuScenes datasets, which test whether the learned dynamics prior transfers beyond internal data. Results are reported separately for each dataset, since aggregation can hide domain-specific failures. This per-dataset view tests whether the intent representation captures transferable vehicle-dynamics structure across different fleets, sensors, and driving styles.

    \subsection{Downstream AEB Evaluation}
    The downstream AEB evaluation treats each intent predictor as a source of future ego trajectories for collision check. The evaluation intentionally abstracts away perception and environment forecasting; at every evaluated timestamp, the future environment state is provided by the dataset and held fixed across all methods. Target agents, static scene structure, and future occupancy/collision geometry are therefore identical for CTRV, AR, and flow models.
    
    For each candidate trajectory, we sweep the ego footprint along the predicted waypoints and check for overlap with the fixed future environment. An AEB trigger occurs when \emph{all} trajectories of the queried candidate set collide. This simplified rule abstracts away production-specific timings, braking, and filtering logic, and focuses the evaluation on whether the predicted ego futures provide useful collision evidence.
    
    We evaluated on two internal AEB datasets. The true-positive set contains 228 NCAP-style  \cite{euroncap_collision_avoidance_2024} collision scenarios and measures whether each trajectory predictor produces ego futures that trigger before the actual collision, penalizing early triggers and rewarding triggers closer to expert driver's intervention. The false-positive evaluation uses approximately 98000 targeted non-collision samples selected to stress-test trajectory-based AEB logic. This set is not a production-level FP estimate; it measures whether an intent distribution can reduce spurious collision triggers relative to a single kinematic extrapolation while preserving strong TP performance.
    
    \section{Results}
    \label{sec:results}
    
    \subsection{Intent Distribution Quality}
    
    Table~\ref{tab:openloop} summarizes the open-loop results, with the best value highlighted for each dataset, metric, and value of \(k\). For \(k\)-CTRV, we keep the speed fixed and roll out curvature hypotheses around the measured curvature, up to \(\pm0.08\,\mathrm{m}^{-1}\). The first candidate is the nominal CTRV path; additional candidates add symmetric left/right curvature offsets. Both AR and flow models reported here are trained on an open-source mix (Waymo, nuScenes) and a Wayve-internal dataset. They consistently outperform the CTRV baseline in all metrics. The high miss-rate for CTRV highlights the fact that driver intent follows a complex distribution and it is insufficient to model it with simple kinematic models.

\newcommand{\oltriplet}[3]{#1/#2/#3}
\begin{table}[b]
    \centering
    \caption{Open-loop trajectory quality. \scriptsize ADE (m) $\downarrow$ / FDE (m) $\downarrow$ / MR (\%) $\downarrow$}
    \label{tab:openloop}
    \footnotesize
    \setlength{\tabcolsep}{2.3pt}
    \renewcommand{\arraystretch}{0.92}
    \begin{tabular}{llccc}
        \toprule
        Method & Dataset & $k=1$ & $k=5$ & $k=10$ \\
        \midrule
        \multirow{3}{1.0cm}{CTRV top-$k$}
        & Wayve    & \oltriplet{2.38}{2.88}{62.4} & \oltriplet{2.30}{2.64}{53.2} & \oltriplet{2.24}{2.45}{46.7} \\
        & Waymo    & \oltriplet{1.54}{1.95}{43.7} & \oltriplet{1.54}{1.94}{43.3} & \oltriplet{1.53}{1.91}{42.4} \\
        & nuScenes & \oltriplet{1.31}{1.77}{38.6} & \oltriplet{1.30}{1.74}{37.8} & \oltriplet{1.30}{1.72}{36.9} \\
        \midrule
        \multirow{3}{1.0cm}{AR OS mix w/ ind.}
        & Wayve    & \oltriplet{0.31}{0.89}{5.8} & \oltriplet{0.20}{0.53}{\textbf{1.5}} & \oltriplet{0.16}{0.43}{\textbf{0.9}} \\
        & Waymo    & \oltriplet{\textbf{0.28}}{\textbf{0.73}}{\textbf{6.3}} & \oltriplet{0.21}{0.51}{\textbf{2.6}} & \oltriplet{0.17}{0.42}{\textbf{1.8}} \\
        & nuScenes & \oltriplet{\textbf{0.44}}{\textbf{1.05}}{\textbf{13.1}} & \oltriplet{\textbf{0.27}}{\textbf{0.60}}{\textbf{3.5}} & \oltriplet{\textbf{0.24}}{\textbf{0.52}}{\textbf{2.5}} \\
        \midrule
        \multirow{3}{1.0cm}{Flow OS mix w/ ind.}
        & Wayve    & \oltriplet{\textbf{0.25}}{\textbf{0.70}}{\textbf{3.6}} & \oltriplet{\textbf{0.18}}{\textbf{0.48}}{1.6} & \oltriplet{\textbf{0.15}}{\textbf{0.37}}{\textbf{0.9}} \\
        & Waymo    & \oltriplet{\textbf{0.28}}{0.74}{6.9} & \oltriplet{\textbf{0.19}}{\textbf{0.47}}{3.0} & \oltriplet{\textbf{0.15}}{\textbf{0.36}}{1.9} \\
        & nuScenes & \oltriplet{0.56}{1.33}{23.0} & \oltriplet{0.35}{0.79}{8.8} & \oltriplet{0.28}{0.59}{4.7} \\
        \bottomrule
    \end{tabular}%
    
\end{table}
    \subsection{AEB as a Downstream Evaluation Task}
    
    Table~\ref{tab:aeb} compares TP and FP behavior on the AEB task. Given the nature of NCAP TP scenarios, the $1$-CTRV trajectory is always on the collision path and scores $100\%$ on TP tests, but is over-sensitive for in-the-wild nominal driving scenarios leading to a very high FP rate. Increasing $k$ improves the FP score, but the resulting trajectories are uniformly spread over a large feasible space and fail to capture a plausible driver intent in safety critical scenarios severely affecting the TP score. The flow model with $k=10$ is our best performing distribution with 25.1$\times$ reduction in FP rate while preserving $98.7\%$ TP score. This indicates that the candidate trajectories sampled from our trained distributions coupled with an all-collide policy does not simply suppress all braking, but also preserves most true-positive behavior enabling a lower false-positive operating point.
    
    \subsection{Ablation  Trends}
    
    Due to space constraints, we omit detailed ablation tables and summarize the main trends. Adding the open-source mix substantially improves cross-domain coverage on Waymo and nuScenes compared with Wayve-only training, while retaining strong Wayve performance. Indicator input does not produce a consistent aggregate gain in the current flow runs, suggesting its effect is scenario-dependent. Varying the history context upto 1.6 s shows that 1.2 s gives the best overall balance across open-loop quality and downstream AEB behavior. Varying waypoint prediction horizons upto 2.0 s, CTRV's performance drops with monotonically widening gap by 25\% vs 1\% for the distribution methods, showing improved horizon scaling.

    \newcommand{\aebcell}[2]{#1/#2}

    \begin{table}[t]
        \centering
        \caption{AEB evaluation. \scriptsize TP-score (\%) $\uparrow$ / FP-rate (\%) $\downarrow$}
        \label{tab:aeb}
        \scriptsize
        \setlength{\tabcolsep}{2.3pt}
        \renewcommand{\arraystretch}{0.92}
        \begin{tabular}{lccc}
            \toprule
            Method & $k=1$ & $k=5$ & $k=10$ \\
            \midrule
            CTRV top-$k$
            & \aebcell{\textbf{100.0}}{1.51}
            & \aebcell{34.7}{0.65}
            & \aebcell{29.5}{0.07} \\
            
            AR OS mix w/ ind.
            & \aebcell{94.9}{2.09}
            & \aebcell{94.9}{0.27}
            & \aebcell{94.9}{0.11} \\
            
            Flow OS mix w/ ind.
            & \aebcell{99.7}{\textbf{0.35}}
            & \aebcell{\textbf{99.0}}{\textbf{0.11}}
            & \aebcell{\textbf{98.7}}{\textbf{0.06}} \\
            \bottomrule
        \end{tabular}
    \end{table}
    
    \section{Conclusion}
    \label{sec:conclusion}
    
    We presented a distributional driver-intent model that predicts multiple plausible future ego trajectories from recent vehicle dynamics alone. However, a blind model cannot reason about unseen objects, traffic lights, lane topology, or map context. We simply showed that large scale expert driving data can expose a useful distribution over driver intent which can be a powerful enabler for downstream tasks. Ranked samples can be queried from such a distribution and leveraged differently depending on the application requirements.
	
	\bibliographystyle{IEEEtran}
	\bibliography{root} 
	
\end{document}